# Mental state attribution to educational robots: an experience with children in primary school

Cristina Gena [1,2], Sara Capecchi [1,2],

[1] *Computer Science Department, University of Turin, Italy*
[2] *Laboratorio Informatica e Scuola, CINI, Italy*

**Abstract**
The work presented in this paper was carried out in the context of the project *Girls and boys: one day at university*[1] promoted by the City of Turin together with the University of Turin. We were responsible for two educational activities on i) robotics and ii) coding hosted at the Computer Science Department, which made one of its laboratories available for this kind of lesson. At the conclusion of the lab's sessions, children compiled the Attribution of Mental State (AMS) questionnaire, which is a measure of mental states that participants attribute to robots, namely the user's perception of the robot's mental qualities as compared to humans. We distributed the questionnaires both to children attending the educational robotics lab and to children performing coding activities. Results show that the first group attributed higher mental qualities to the robots, compared to the attribution given by children that did not have a direct experience with a robot.

## 1. Introduction

The mental state attribution has been defined as "the cognitive capacity to reflect upon one's own and other persons' mental states such as beliefs, desires, feelings and intentions" [1]. In everyday human-to-human interactions, such attributions are ubiquitous, although we are typically not necessarily aware of the fact that they are attributions—or the fact that they are attributions of mental states.

According to Thellman et al. [21], the mental state attribution to robots is a complex socio-cognitive process. Despite a widespread belief that robots do not have minds [17], people frequently talk about and interact with robots as if they have minds. A common conception is that mental state attribution helps people interact with robots by providing an interpretative framework for predicting and explaining robot behavior [9].

People's tendency to attribute mental states to robots is determined by multiple factors, including age and motivation as well as robot behavior, appearance, and identity, etc. even if there is not a clear consensus about the reasons why people attribute mental states to robots [21]. A total of 31 studies reviewed by Thellman et al. [21] indicate that robot behavior determines the tendency to attribute mental states to robots. There is corroborating evidence that people are more inclined to attribute mental states to robots when they exhibit various types of socially interactive behavior, such as eye gaze, gestures, cheating, emotional expression, and when behavior is unpredictable, complex, intelligent, or highly variable. Also, the definition of a robot personality plays an important role in the children's attribution of mental state to robots [5][10].

---

[1]https://www.unito.it/ateneo/gli-speciali/bambine-e-bambini-un-giorno-alluniversita

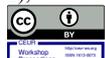

There is evidence of a stronger tendency in children (particularly young children) in the attribution of mental states to robots compared with adults [6][7][14][15][16]. It should be noted that most of the studies reporting these findings employed verbal measures of mental state attribution [21], as for instance Likert or semantic differential scale.

According to the review by Thellman et al. [21], a typical study on mental state attribution to robots is conducted in a lab setting, is based on WEIRD participants (i.e., Well-Educated, Industrialized, Rich, and Democratic), presents study participants with a representation of a robot (e.g., image or text) as a stimulus materials presented to participants, and employs a verbal measure, probably Likert or semantic differential scale, to study one of its determinants. The predominant tool used in studies based on a child population is spoken or written questions about the mental states of robots in combination with a binary choice response format (i.e., typically yes-no questions). When the study presents a robot in presence, it usually exhibits social behavior in the context of a direct interaction with the study participant.

In the framework of both a direct and indirect interaction with an educational robot we carried out a study on attribution of mental state to robots. The work presented in this paper was carried out in the context of the initiative called "*Girls and boys: one day at university*" promoted by the City of Turin together with the University of Turin. We were responsible for the educational robotics [4][11], unplugged [2], and coding activities carried out at the Computer Science Department, which made one of its laboratories available for this kind of lesson. At the conclusion of two lab's sessions, children compiled the Attribution of Mental State (AMS) questionnaire [6][7][14], which is a measure of mental states that participants attribute to robots, and in this context is the children's perception of the robot's mental qualities as compared to humans. We distributed the questionnaires both to children attending the educational robotics lab and to children performing coding activities. Results show that children that interacted with the robots attributed higher mental qualities to the robot, compared to the attribution given by children that did not have a direct experience with a robot.

## 2. The context and the project
## 2.1. The project *Girls and boys: one day at university*

The project *Girls and boys: one day at university* is a public engagement initiative promoted by the City of Turin and ITER - Turin Institution for Responsible Education within the "Growing Up in the City" project and involving the University of Turin. The initiative proposes workshops, guided tours and educational paths designed specifically for students of primary and lower secondary schools with the aim of building an imaginary access to higher education in conditions of equal opportunities, and to disseminate among the very young students the impact of the research in our daily lives. At the same time, the project is an opportunity for teachers, researchers, and PhD students to enhance their commitment in the field of disseminating the results of their work and to experiment with new languages and methods for communicating this knowledge. The Computer Science department, and in particular the k-12 Education research group, took part in this project for many years, organizing workshops and lessons to promote the so-called *computational thinking* [12][22] through unplugged activities [2], coding[3] and educational robotics lessons [4][11].

## 2.2. The educational robotics tutorial

In 2022 we designed and then carried out an educational robotics tutorial, organized as the worldwide famous *One Hour of Code*[4], which is a one-hour introduction to computer science, using fun tutorials, typically delivered through coding exercises such as the ones based on Google Blocky or Scratch, to show that anybody can learn the basics of coding and programming. This campaign is supported by over 400 partners and 200,000 educators worldwide.

---
[3] https://programmailfuturo.it/

[4] https://hourofcode.com/

Our educational robotic tutorial was based on the mBot robot[5] and its graphical block editor called mBlock[6] (see Fig. 1), installed on a tablet. mBot is a small robot, base d on Arduino, the famous open-source platform that allows users and designers to create small-sized devices. mBlock is a programming environment developed by Makeblock[7] that gives the possibility to work both locally, by installing software (as was done for children) or to use the IDE present on the official page via a browser.

In order to introduce primary school's children to the basic of coding and robotics, we design a tutorial for mBot organized as a two-hours tutorial as follows:

- introduction to the robot and its features as sensors, leds, movements,
- introduction to mBlock and its features, as stage area, blocks area and script area,
- introduction and tutorial on the basic robot movements,
- introduction to the led panel and exercise on how to make a drawing appear on it,
- introduction on the ultrasonic sensors and exercise on obstacle avoidance. This exercise is an opportunity to introduce the conditional construct if, that is, for example, *if mBot sees an obstacle less than 20 cm away… something happens*.

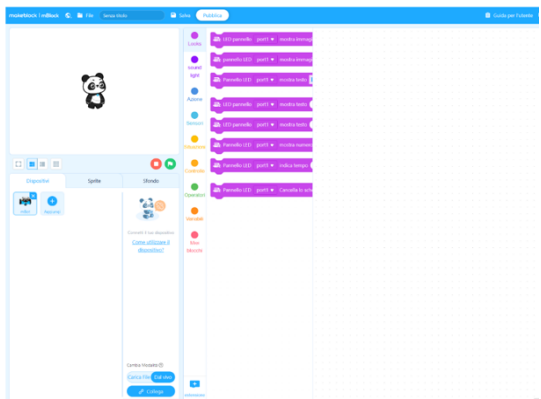

**Figure 1**: The mBlock environment

## 2.3. The coding tutorial

---

[5] https://www.makeblock.com/steam-kits/mbot
[6] https://mblock.makeblock.com/en-us/
[7] https://www.makeblock.com/

The coding tutorial was taken from the Minecraft Hour of Code tutorial on a tablet PC. The tutorial is available on code.org and works well for any students old enough to read, but with younger learners trying hard to finish the tutorial, and older students having some time to play on the free play level at the end. The lesson faces topics as sequences, cycles, events, and conditions (at the very end, for those students who completed all the steps).

## 3. Investigation on mental state attribution to educational robots
## 3.1. Methodology

At the end of both the activities, we distributed a survey to the children, to collect information about their satisfaction and engagement. The survey consisted of 15 questions (detailed in the Appendix) in which children can respond by giving assessments included in a scale of five values expressed thanks to a Smileyometer [18], which is the most used tool for the measurement of children's opinion and includes an evaluation scale [3] through the smileys corresponding to a range from 1 to 5 (from strongly disagree to strongly agree). The children were asked to express their opinion by choosing one of the faces. The survey was anonymous, to protect children's privacy[8].

After the survey, children also compiled the Attribution of Mental State (AMS) questionnaire [6][7], which is a measure of mental states that participants attribute to robots, is the children's perception of the robot's mental qualities as compared to humans. The AMS consists of five dimensions: *Epistemic, Emotional, Desires and Intention, Imaginative, and Perceptive*. The epistemic dimension concerned the participants' idea of the robot's *cognitive intelligence* (e.g., *Can the robot understand/decide/learn/teach/think?*), the perceptive dimension is related to the possible robot perception and sensation (e.g., *Can the robot smell/watch/taste/listen/feel cold?*). The other dimensions concerned the user's mental attribution to the robot's *emotional intelligence,* example questions are: *Can the robot get angry/be scared/be happy?* (Emotional dimension); *might the robot want to do something/make a*

---

[8] The studies reported in this paper were conducted in accordance with the Declaration of Helsinki, and the protocol was approved by the Ethics Committee of the University of Turin (Prot. N. 0596407)

*wish/prefer one thing over another*? (Desires and intention dimension); *can the robot image/tell a lie/make a dream/make a joke?* (Imaginative dimension). The questionnaire consists of 25 questions in which children can respond *a lot, a few, or no*. We decided that children could give their assessments on a scale of three values expressed thanks to the Smileyometer [19]. We included an evaluation scale through the smileys corresponding to the three values: a *lot*, *a little*, *no*, which correspond to the scores 2,1 and 0 respectively. The children were asked to express their opinion by choosing one of the three faces. The user total score is the sum of all answers (range = 0-50); the five partial scores are the sum of the answers within each dimension (range = 0-10).

### 3.2 Educational robotics tutorial

The class involved in the educational robotics activity was made up of 24 children (11 males, 13 females). The children were divided into pairs and equipped with an Android tablet with the mBlock IDE preloaded, and with a mBot robot for each pair. The tutorial lasted 2 hours and was given by one professor of the Department assisted by 3 master students.

Even if the children found the lesson interesting and enjoyed themselves, they were not able to conclude all the topics set for the two available hours. There have been some slowdowns mostly caused by the presence of the robot which gives rise to that interest and curiosity in wanting to try as much as possible. For this reason, children focused more on the practical part by trying the various exercises rather than on the theory part. Often the children, in addition to the pre-structured exercises, were able to create other types of actions with the robot, even using blocks not previously introduced. This shows that, on the one hand, the block environment is clear to understand, and, on the other hand, the type of lesson makes the children want to experiment and try.

The feedback obtained by the satisfaction and engagement survey allowed us to conclude that the lessons have been appreciated and have produced satisfactory results; more than 68% of the judgments obtained the strongly agree level of satisfaction and 17% obtained an agree level, with 75% of children that strongly agreed on *I am satisfied with my results and progress* and a 95% that strongly agreed that they would like to *Have more lessons of this type*.

Concerning the Attribution of Mental State (AMS) questionnaire we analyzed the partial and total scores, and ranked the dimensions based on its average values:
- The epistemic dimension received an average value of 1.07 (SD=0.18)
- The desires and intention dimension received an average value of 1.05 (SD=0.19)
- The perceptive dimension received an average value of 1.04 (SD=0.59)
- The emotional dimension received an average value of 1.01 (SD=0.18)
- The imaginative dimension received an average value of 0.69 (SD=0.23)

The results showed that, overall, the mental states attributed to the robot received on average a medium score. It is interesting to note that children attributed higher scores to the cognitive dimension than others, as for instance emotion and imagination received a lower score. Therefore, children attributed knowledge and intentionality to the robot rather than attributions related to the emotional sphere. The good positioning of perceptual abilities may have been influenced by the presence of the robot's sensors.

Looking at the frequency distribution of *a lot, a little, no* answers, they respectively received 141, 147, 149 answers, witnessing a quite well balanced situation, with the following question/answer receiving higher scores: *the robot can understand* (a lot: 12 out of 21 answers, 57%), *the robot can be happy* (a lot: 11 out of 21 answers, 57%), *the robot can look* (a lot: 11 out of 15 answers, 73%), *the robot can listen* (a lot: 11 out of 14 answers, 78%).

### 3.2 One hour of code tutorial

After having analyzed the above results about the mental state attribution to the educational robot mBot, we compared them with the ones coming from similar students (i.e., same class level) who had no experience with an educational robot. Therefore, after one of our coding lessons given in the context of the project *Girls and boys: one day at university*, children also compiled the AMS questionnaire.

After the coding activities children compiled the satisfaction survey, and the AMS questionnaire, showing, at the beginning, a brief description, and pictures of the mBot robot. Results were very different, as detailed in the following.

After having collected and analyzed the results, we summed up all the partial scores and ranked the dimensions based on their average values:
- The perceptive dimension received an average value of 0.58 (SD=0.55)
- The epistemic dimension received an average value of 0.48 (SD=0.26)
- The desires and intention dimension received an average value of 0.46 (SD=0.32)
- The emotional dimension received an average value of 0.35 (SD=0.21)
- The imaginative dimension received an average value of 0.23 (SD=0.06)

Again, the perceptive and the epistemic dimensions are the most selected, however results show scores are very low, witnessing the fact that children with no direct experience with the educational robot probably tend to not attribute mental states to the robot. This is also confirmed by the distribution analysis: looking at the frequency distribution of *a lot, a little, no* answers, they respectively received 66, 91, 370 answers, with the *no option* chosen in most cases. Scores are lower than the ones obtained in the educational robotics experience, and statistically analyzing the differences between the AMS scores in the two groups, we found them to be significant, as witnessed by a paired t-test, p=0.0109.

## 4. Conclusion and future work

Several researchers in the field [21] suggested that, in studying the mental attribution to robots, it is not just useful but also necessary to conduct studies *in the wild* and eventually compare results with those from the lab. We performed an evaluation in a real context and in two separate situations. Results show that children that approached an educational robot attributed higher mental qualities to the robot itself, compared to the attribution given by children that did not have a direct experience with a robot. These findings are interesting but preliminary and need to be confirmed in other evaluations, and possibly in longitudinal studies, and in pre- and post-test measuring the changes of mental state attribution before and after the direct experience with a robot. Our study also suggests that research in which the involved users have not had direct experience with robots, could present results that tend to be different from those who have had a different experience, and which could change if users could interact directly with a robot.

## 5. References

[1] Martin Brüne, Mona Abdel-Hamid, Caroline Lehmkämper, and Claudia Sonntag. 2007. Mental state attribution, neurocognitive functioning, and psychopathology: What predicts poor social competence in schizophrenia best? *Schizophrenia Research* 92, 1–3 (2007), 151–159.

[2] Sara Capecchi, Cristina Gena, and Ilaria Lombardi. 2022. Visual and unplugged coding with smart toys. In Proceedings of the 2022 International Conference on Advanced Visual Interfaces (AVI 2022). Association for Computing Machinery, New York, NY, USA, Article 26, 1–5. https://doi.org/10.1145/3531073.3531180

[3] Federica Cena, Cristina Gena, Pierluigi Grillo, Tsvi Kuflik, Fabiana Vernero, Alan J. Wecker: How scales influence user rating behaviour in recommender systems. Behav. Inf. Technol. 36(10): 985-1004 (2017)

[4] Valerio Cietto, Cristina Gena, Ilaria Lombardi, Claudio Mattutino, Chiara Vaudano: Co-designing with kids an educational robot. ARSO 2018: 139-140

[5] Rossana Damiano, Cristina Gena, Andrea Maieli, Claudio Mattutino, Alessandro Mazzei, Elisabetta Miraglio, Giulia Ricciardiello: UX Personas for defining robot's character and personality. CoRR abs/2203.04431 (2022)

[6] Di Dio, C., Isernia, S., Ceolaro, C., Marchetti, A., & Massaro, D. (2018). Growing Up Thinking of God's Beliefs: Theory of Mind and Ontological Knowledge. SAGE Open, 8(4). https://doi.org/10.1177/2158244018809874

[7] Di Dio, C., Manzi, F., Itakura, S. et al. It Does Not Matter Who You Are: Fairness in Preschoolers Interacting with Human and Robotic Partners. Int J of Soc Robotics 12, 1045–1059 (2020). https://doi.org/10.1007/s12369-019-00528-9

[8] Duque, I.; Dautenhahn, K.; Kheng Lee Koay; Willcock, L.; Christianson, B., "A different approach of using Personas in human-robot interaction: Integrating Personas as computational models to modify robot companions' behaviour," RO-MAN, 2013 IEEE, vol., no., pp.424,429 (2013)

[9] Nicholas Epley, Adam Waytz, and John T. Cacioppo. 2007. On seeing human: A three-factor theory of anthropomorphism. *Psychological Review* 114, 4 (2007), 864–886.

[10] Cristina Gena, Claudio Mattutino, Andrea Maieli, Elisabetta Miraglio, Giulia Ricciardiello, Rossana Damiano, Alessandro Mazzei: Autistic Children's Mental Model of an Humanoid Robot. UMAP (Adjunct Publication) 2021: 128-129

[11] Cristina Gena, Claudio Mattutino, Gianluca Perosino, Massimo Trainito, Chiara Vaudano, and Davide Cellie. Design and development of a social, educational and affective robot. In *2020 IEEE Conference on Evolving and Adaptive Intelligent Systems, EAIS 2020, Bari, Italy, May 27-29, 2020*, pages 1–8. IEEE, 2020.

[12] Peter B. Henderson, Thomas J. Cortina, and Jeannette M. Wing. Computational thinking. In *Proceedings of the 38th SIGCSE Technical Symposium on Computer Science Education*, SIGCSE '07, page 195–196, New York, NY, USA, 2007. Association for Computing Machinery.

[13] Anthony Jameson, Silvia Gabrielli, Per Ola Kristensson, Katharina Reinecke, Federica Cena, Cristina Gena, and Fabiana Vernero. How can we support users' preferential choice? In *CHI '11 Extended Abstracts on Human Factors in Computing Systems*, CHI EA '11, page 409–418, New York, NY, USA, 2011. Association for Computing Machinery.

[14] Manzi, F., Di Dio, C., Itakura, S., Kanda, T., Ishiguro, H., Massaro, D. et al. (2020). Moral evaluation of Human and Robot interactions in Japanese preschoolers. In cAESAR 2020, ACM IUI Workshop Proceedings. Cagliari, Italy.

[15] Mako Okanda, Kosuke Taniguchi, Ying Wang, and Shoji Itakura. 2021. Preschoolers' and adults' animism tendencies toward a humanoid robot. *Computers in Human Behavior* 118 (2021), 106688.

[16] Sandra Y. Okita, Daniel L. Schwartz, Takanori Shibata, and Hideyuki Tokuda. 2005. Exploring young children's attributions through entertainment robots. In Proceedings of the ROMAN 2005. IEEE International Workshop on Robot and Human Interactive Communication, 2005. IEEE, 390–395.

[17] Ceylan Özdem, Eva Wiese, Agnieszka Wykowska, Hermann Müller, Marcel Brass, and Frank Van Overwalle. 2017. Believing androids–fMRI activation in the right temporo-parietal junction is modulated by ascribing intentions to non-human agents. *Social Neuroscience* 12, 5 (2017), 582–593.

[18] Rachel L. Severson and Kristi M. Lemm. 2016. Kids see human too: Adapting an individual differences measure of anthropomorphism for a child sample. Journal of Cognition and Development 17, 1 (2016), 122–141.

[19] Gavin Sim and Matthew Horton. 2012. Investigating children's opinions of games: Fun Toolkit vs. This or That. In Proceedings of the 11th International Conference on Interaction Design and Children (IDC '12). ACM, New York, NY, USA, 70-77.

[20] Mark C. Somanader, Megan M. Saylor, and Daniel T. Levin. 2011. Remote control and children's understanding of robots. *Journal of Experimental Child Psychology* 109, 2 (2011), 239–247.

[21] Sam Thellman, Maartje de Graaf, and Tom Ziemke. 2022. Mental State Attribution to Robots: A Systematic Review of Conceptions, Methods, and Findings. J. Hum.-Robot Interact. 11, 4, Article 41 (December 2022), 51 pages. https://doi.org/10.1145/3526112

[22] Lodi, M., Martini, S. Computational Thinking, Between Papert and Wing. Sci & Educ 30, 883–908 (2021). https://doi.org/10.1007/s11191-021-00202-5

## Appendix

In the following we listed the survey questions distributed to the children

1. I had already had experience with robots before these lessons
2. I find the activities offered to me interesting
3. I find the activities that are proposed to me easy
4. In the face of difficulties, I increase my commitment
5. I can perform a task by myself
6. I try to learn from my mistakes
7. I acquired new notions in coding
8. It's easy to remember what I studied
9. I can concentrate during the lesson
10. Time passes quickly during these lessons
11. I get help when I'm in trouble

12. Outside of school I use at least one of these means: pc, tablet, robot, smartphone.
13. When I go home, I am satisfied with the experiences I had at school
14. I am satisfied with my results and progress
15. I wish I had more lessons like this